\documentclass[letterpaper, 10 pt, conference]{ieeeconf}  % Comment this line out if you need a4paper

\IEEEoverridecommandlockouts                              % This command is only needed if 
\usepackage{graphicx} % for pdf, bitmapped graphics files
\usepackage[utf8]{inputenc}
\usepackage{amsmath,amsfonts,booktabs,cite} % general useful stuff
\usepackage{gensymb} % for \degree
\usepackage{textcomp} % for the degree symbol
\usepackage[separate-uncertainty=true]{siunitx}%
\usepackage[hidelinks]{hyperref} % to have hyperlinks and for the \ref macros
\usepackage{cleveref} % for easier referencing
\usepackage{tabularx}
\usepackage{pbox}
\let\labelindent\relax
\usepackage[inline]{enumitem} % inline enumerate
\usepackage[nolist]{acronym} % acronyms by the \ac{label} command
\usepackage{multirow}
\usepackage[linesnumbered,ruled,vlined]{algorithm2e}
\usepackage{subfigure}
\usepackage{algpseudocode}
\usepackage{indentfirst}
\usepackage[font=small]{caption}

\def\secref#1{Sec.~\ref{#1}}

\def\eqref#1{Eq.~(\ref{#1})}

\newcommand\etal{~\emph{et al. }}

\newsavebox{\twosubbox}

\crefname{algocf}{alg.}{algs.}
\Crefname{algocf}{Algorithm}{Algorithms}

\title{\LARGE \bf
  Subgoal-Driven Navigation in Dynamic Environments Using \\Attention-Based Deep Reinforcement Learning
}

\author{Jorge de Heuvel \and Weixian Shi \and Xiangyu Zeng \and Maren Bennewitz% <-this % stops a space
\thanks{All authors are with the Humanoid Robots Lab, University of
  Bonn, Germany. Maren Bennewitz is additionally with the Lamarr
  Institute for Machine Learning and Artificial Intelligence,
  Germany. This work has partially been funded by the Deutsche Forschungsgemeinschaft (DFG, German Research Foundation) under the grant number BE~4420/2-2~(FOR 2535 Anticipating Human Behavior).}}

\begin{document}

\maketitle
\thispagestyle{empty} 
\pagestyle{empty}

\begin{abstract} Collision-free, goal-directed navigation in environments containing unknown static and dynamic obstacles is still a great challenge, especially when manual tuning of navigation policies or costly motion prediction needs to be avoided.
In this paper, we therefore propose a subgoal-driven hierarchical navigation architecture that is trained with deep reinforcement learning and decouples obstacle avoidance and motor control.
In particular, we separate the navigation task into the prediction of the next subgoal position for avoiding collisions while moving toward the final target position, and the prediction of the robot's velocity controls.
By relying on 2D lidar, our method learns to avoid obstacles while still achieving goal-directed behavior as well as to generate low-level velocity control commands to reach the subgoals.
In our architecture, we apply the attention mechanism on the robot's 2D lidar readings and compute the importance of lidar scan segments for avoiding collisions.
As we show in simulated and real-world experiments with a Turtlebot robot, our proposed method leads to smooth and safe trajectories among humans and significantly outperforms a state-of-the-art approach in terms of success rate.
A supplemental video describing our approach is available online\footnote{Full video:
	\href{https://www.hrl.uni-bonn.de/publications/deheuvel23iros_subgoal.mp4}{hrl.uni-bonn.de/publications/deheuvel23iros\_subgoal.mp4}.
} 
\end{abstract} 

\section{Introduction}
\label{sec:intro}
% DONE

Classical navigation methods, such as extensions of the popular Dynamic Window Approach~\cite{Missura22iros,dwa} or Reciprocal Collision Avoidance~\cite{orca} compute motion commands for vehicles to move collision-free towards the goal by assuming a constant velocity of moving obstacles.
However, due to the lack of knowledge about their future motions, these methods can encounter problems in case of dynamic obstacles such as humans.
Typically, one needs to carefully design and tune these methods to be able to deal with the different situations that include dynamic obstacles of different number, direction, and velocity.

To avoid tedious tuning of parameters, learning-based methods have gained popularity in the recent years.
By realizing obstacle avoidance and or even the entire navigation pipeline with deep reinforcement learning (RL), collision-free navigation behavior can be learned in a sophisticated simulation environment.
Through the use of neural networks (NN) in the learning process, navigation relevant features can be implicitly identified from various sensor setups.
Recently, successful approaches have used attention-based techniques or graph learning to encode the relationships between the robot and surrounding humans~\cite{liu2021decentralized, chen2019crowd, chen2020relational}.
These and other methods achieve good performance in dynamic crowd scenarios under the assumption of full observability of the moving obstacles including their velocity~\cite{everett2018motion,wheretogo}.
However, full observability is a strong assumption to make, as it is difficult to robustly estimate all human dynamics in real-world conditions.

\begin{figure}[t]
	\centering
	\includegraphics[width=\linewidth]{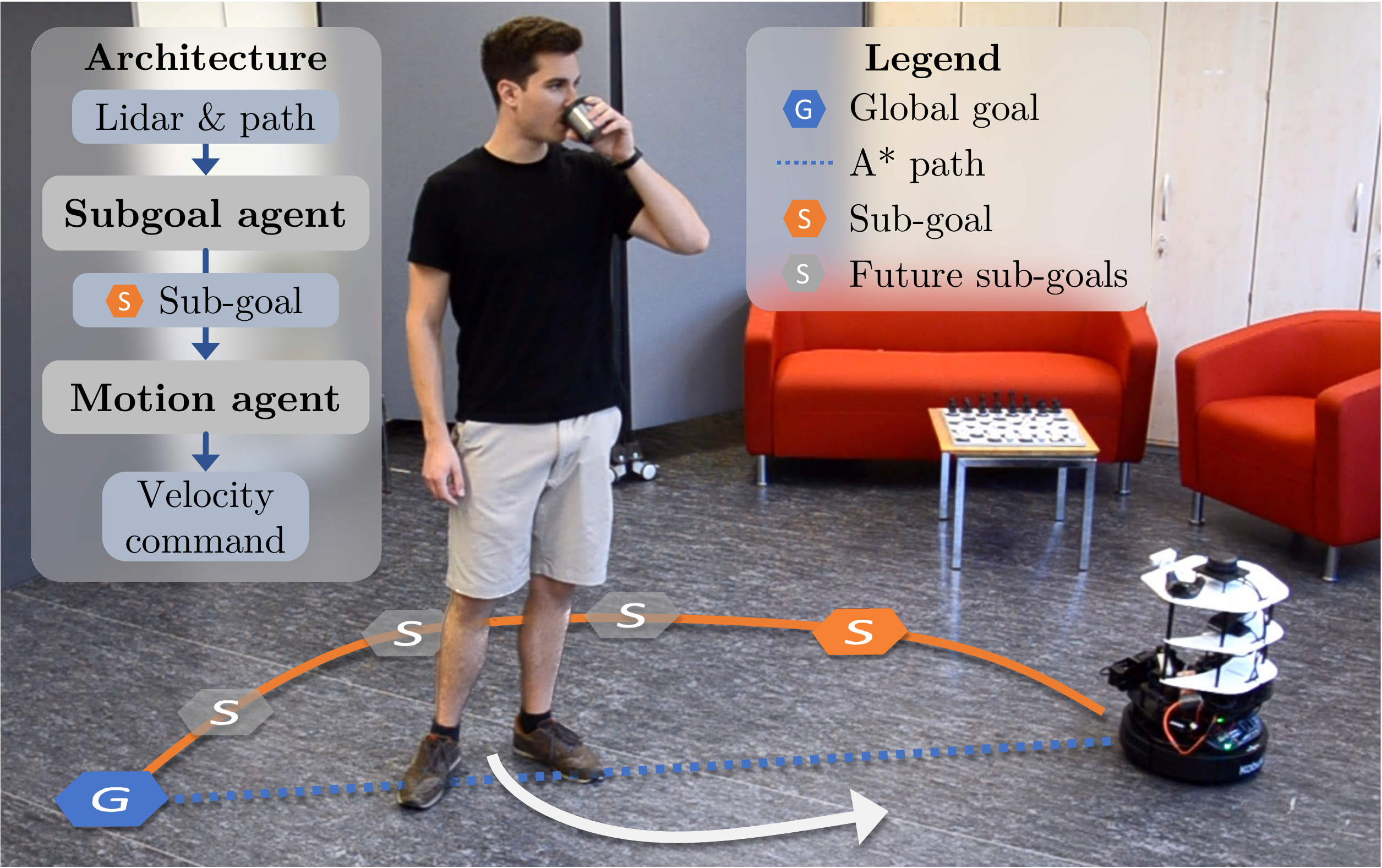}
	\caption{
		In our hierarchical learning-based navigation framework, the subgoal agent predicts the next subgoal position to avoid possible collisions with unknown static or dynamic objects.
		It does this using only the global path to the goal and lidar observations, which are spatially processed in an attention-based architecture.
		The robot then moves toward the nearby subgoal position under the control of a low-level motion agent.
	} 
%	\vspace{-2ex}
	\label{fig:motivation}
\end{figure}

In this paper, we propose a subgoal-driven hierarchical reinforcement learning framework based on lidar observations.
It is designed to deal with unknown dynamic and static obstacles that a robot may encounter on its way to a global goal, while no knowledge of the obstacle dynamics is required.
Our approach features two hierarchical levels.
The high level \textit{subgoal agent} provides cleverly placed subgoal positions that aim at avoiding collisions along the way to the target location.
At the low level, our \textit{motion agent} subsequently generates the velocity controls for the robot to pursue the provided subgoal.
By applying this two-level architecture, the tasks of obstacle avoidance and velocity control are decoupled, allowing the subgoal agent to focus on collision-free, goal-directed navigation without having to output actual velocity commands.
In our architecture, we implemented the attention mechanism to compute the spatial importance lidar scan segments.
As we show in our experiments with a Turtlebot robot navigating in dynamic indoor environments, our attention-based, subgoal-driven navigation framework significantly outperforms the navigation policy generated by a strongly related baseline approach and enables safe navigation in indoor environments with humans in motion.

To summarize, the main contributions of our work are: \begin{itemize} 
	\item A hierarchical navigation architecture for obstacle avoidance in dynamic indoor environment based on subgoals.
	\item Applying the attention mechanism on lidar measurements to weigh their spatial importance and extract essential features for collision avoidance.
	\item An experimental evaluation demonstrating the significantly higher success rate compared to a state-of-art approach~\cite{perez2021robot}.
	\item Transferring the learned policy smoothly to a real robot (see video attachment).
\end{itemize} 

\section{Related Work}
% =======================================
% Learning Navigation - DONE
% =======================================
Several learning-based navigation methods have been presented that show great performance in environments containing static and dynamic obstacles \cite{everett2018motion, chen2019crowd, chen2020relational, liu2021decentralized, perez2021robot, shienhanced, de_heuvel_learning_2022, dawood_handling_2022, chen_robot_2020}.
Among them are learning frameworks focusing specifically on crowd~\cite{liu2021decentralized, perez2021robot} or personalized navigation in the user's vicinity~\cite{de_heuvel_learning_2022, de_heuvel_learning_2022-1}.
% Graph:
To improve the performance of crowd navigation, Chen\etal\cite{chen2020relational} encoded the robot-human and human-human relationships as a relational graph.
\mbox{Liu\etal\cite{liu2021decentralized}} introduced a spatial-temporal graph through a structural recurrent NN to handle the complex dynamic navigation problem.
% Observability of velocities
However, most of the approaches assume that obstacle dynamics are known or can easily be inferred, which is not feasible in real-world environments.

% =======================================
% Attention Mechanism - DONE
% =======================================
Inspired by the significant success of transformer networks for language models, the use of attention mechanisms with deep reinforcement learning is a promising avenue.
For robot navigation in crowded environments under the assumption of full observability, attention mechanisms have lead to improved performance compared other network architectures~\cite{chen_crowd-robot_2019, shi_enhanced_2022}.
Also, sensor fusion via attention networks was successfully realized by Weerakoon\etal~\cite{weerakoon_sim--real_2022} for costmap-based navigation in uneven outdoor environments.
Han\etal\cite{han_deep_2022} fused depth and lidar sensor data with attention networks for indoor navigation.
Based on these findings, we will leverage the attention mechanism in our approach to learn dynamic collision avoidance from sensor data directly.

% =======================================
% LIDAR-Based Navigation - DONE
% =======================================
%Regier\etal\cite{regier2020deep} adapted the lidar measurements into a grid local map. 
Lidar sensors provide reliable measurements of surrounding obstacles, which makes them an favorable choice for collision avoidance algorithms~\cite{jia_2d_2022, ryu_confidence-based_2022}.
For example, Surmann\etal\cite{surmann_deep_2020} proposed a novel method of map-less robot navigation with a robust sim-to-real transfer based on a 2D laser sensor and orientation towards the goal in the state space.
P{\'e}rez-D’Arpino\etal\cite{perez2021robot} pass lidar data and waypoints extracted from a global path to convolutional networks and a fully-connected network respectively, which ultimately outputs velocity commands for collision-free navigation.
We follow these authors and choose 2D lidar as our main sensor for obstacle detection and utilize the global path for guidance.
In contrast to~\cite{perez2021robot} however, our framework decouples collision avoidance from motor control and predicts collision-free subgoals.
Note that \cite{perez2021robot} will be employed as a baseline approach in this work.

% =======================================
% Subgoal Navigation - DONE
% =======================================
Predicting subgoals for collision-free robot navigation has been a successful approach in the areas of map-less navigation~\cite{gebauer_sensor-based_2023, zhang_generating_2020, xu_hierarchical_2021}, or collision avoidance with pedestrians~\cite{ah_sen_human-aware_2022, wheretogo, kastner_connecting_2021-1}.
An intrinsic property of many subgoal navigation approaches is their hierarchical nature: As one high-level agent predicts subgoals, a low-level motion controller is needed to navigate towards the subgoal position.
In the approach of Brito\etal\cite{wheretogo}, a subgoal-generating RL agent provides local subgoals to a low-level model predictive control planner for improved navigation in crowded environments.
However, their approach requires knowledge of all surrounding pedestrian poses and dynamics.
To overcome this limitation, we will learn the pedestrian dynamics implicitly from lidar observations and replace the low-level controller with a lightweight motion planning RL agent.

\section{Our Approach}
\label{sec:approach}

\subsection{Overview \& Assumptions}

\begin{figure*}[t] 	\centering 	\includegraphics[width=\linewidth]{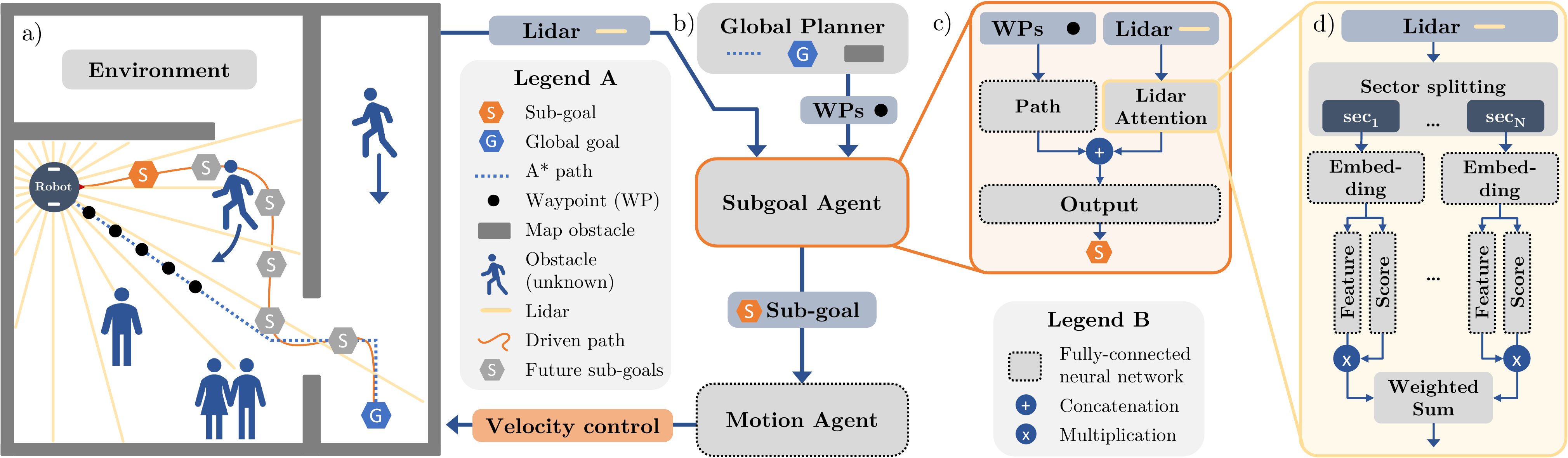} 	\caption{ 
		% DONE
		Schematic of the architecture of our approach: 		\textbf{a)} 		A robot pursuing a global goal encounters unknown dynamic obstacles such as pedestrians on its A* path.
		Local collision avoidance is achieved by predicting and pursuing local subgoals in the vicinity of the robot.
		\textbf{b)} 		Our hierarchical sub-goal navigation approach relies on 2D lidar and 5 upcoming waypoints as input to the subgoal agent (orange).
		The predicted subgoal is passed on to the motion agent, which controls the robot.
		\textbf{c)} 		The subgoal agent consists of a path processing NN (neural network) and an attention module, which encodes the spatial information of the 2D lidar.
		Their respective output is concatenated and further processed in a NN.
		\textbf{d)} 		The attention block (yellow) first splits the lidar measurements into $N=10$ sectors.
		Subsequently encoded via an embedding NN, the sectors are passed into feature and score extracting networks.
		Finally, the scores multiply the features in the weighted sum.
	} 
	\label{fig:architecture}
\end{figure*} 

% DONE
Our goal is to learn a robot navigation policy capable of local obstacle avoidance among unknown static and dynamic obstacles via reinforcement learning, while relying only on a global path and 2D lidar readings as input, compare Fig.~\ref{fig:architecture}a).
The global path is planned using A* in a map containing the known static obstacles of the environment.
The dynamics of the moving obstacles are unknown to the robot and are implicitly learned from subsequent lidar readings for collision-free navigation.

We adapt a hierarchical two-layer framework to tackle the navigation problem, compare Fig.~\ref{fig:architecture}b): The top layer, called \textit{subgoal agent}, predicts a nearby subgoal to the robot.
Pursuing the cleverly placed subgoal prevents collisions with unknown dynamic and static obstacles.
The bottom layer of the framework, called \textit{motion agent}, is driving the robot towards to subgoal via velocity commands.
It is trained to reach the given subgoal as efficiently possible.
Hence, collision avoidance solely is the duty of the subgoal agent.

%\subsubsection{Replanning of A*}
% DONE
After larger deviations from the original global path have taken place in favor of collision avoidance, it can be wise to replan the A* path to get the newest shortest path to the goal~\cite{missura_fast-replanning_2022}.
We adapt this concept and replan the robot's global path every three subgoal predictions.
Note that during replanning, the A* path still does not consider the unknown obstacles in the scene.
If it did consider dynamic obstacles, the path would occasionally be counterproductive as it could fall in the direct motion direction of dynamic obstacles.

\subsection{Subgoal Agent}
% DONE
The objective of the subgoal agent (SA) is to predict a subgoal pursued by the motion agent to avoid collision, while following the global path to the target position.
We use a fixed time frame of $\Delta t_{\mathit{SA}} = 0.2 \text{ }\si{\second}$ between subsequent states of the subgoal agent.
Even though the subgoal agent does not observe a temporal lidar sequence, the fixed time frame of observations benefits the implicit association of a dynamic obstacle's shape in the lidar data with their expected velocity.
The timing is independent of whether a subgoal is reached within $\Delta t_{\mathit{SA}}$.
In this case, the robot would come to a stop at the subgoal, waiting for the next subgoal prediction.

\subsubsection{State Space} 
\label{sec:subgoal_state_space}
% DONE
The state space of the subgoal agent consists both lidar measurements and five upcoming waypoints, as depicted in Fig.~\ref{fig:architecture}a).
The waypoints are sampled along the global A* path to the target location.
Specifically, we determine the closest waypoint on the A* path from the robot's current position, and onward-interpolate four subsequent waypoints with a distance of $0.3\text{ }\si{\meter}$.
Consequently, the five waypoint set covers a distance of $1.2\text{ }\si{\meter}$.
%Given the robot is closest to the $m$-th waypoint, then the considered waypoints at time $t$\todo{brauchen wir t?} is $s_t^w=\left\{(x_{m+i}^w,y_{m+i}^w)|i\in[1,5]\right\}$ \todo{erster waypoint m oder m+1?}. 

The $360\degree$ scanning 2D lidar we use in our experiments outputs 1,440 beams with a maximum range of $12\text{ }\si{\meter}$.
As pre-processing, the $0.25\degree$ resolution of the lidar is down-sampled via min-pooling to $80$ rays at a $4.5\degree$ resolution.
Furthermore, we clip the lidar range at $4 \text{ }\si{\meter}$, as information about obstacles within this range around the robot is sufficient for local obstacle avoidance.
%Furthermore, we divide the range scan into 80 regions and take the beam with the minimum distance.
%Thus, we get 80~distance readings\todo{was wird  mit readings gemacht die laenger als 4m?} denoted as $s_t^l=\left\{(x_i^l,y_i^l)|i\in[0,79]\right\}$. \todo{brauchen wir Index t?}
%
%So the combined state space is defined as
%\begin{eqnarray}
%	\label{eq:subgoal_obs}
%	s_{\text{\textit{SA}}} = (s_l,s_w).
%\end{eqnarray}   
Note that all coordinates in the state space are converted to robot-centric Cartesian 2D coordinates.
% where the robot's position is the origin of the coordinates, and its orientation points to the positive direction of the x-axis.

\subsubsection{Action Space} 
% DONE
The subgoal agent's action space is continuous and consists of the robot-centric subgoal-position in polar coordinates $a_{\text{\textit{SA}}} = (l,\theta)$, 
%\begin{eqnarray}
%	\label{eq:subgoal_action}
%	a_{\text{\textit{SA}}} = (l,\theta)
%\end{eqnarray}   
where $l \in [0,0.6m]$ and $\theta \in [0,2\pi]$.
%The maximum value of $l$ is because we take the second waypoint which is around 0.6m away as a guidance point to compute the reward, see next section\todo{check wether this makes sense}.
Note the edge case, where by choosing a subgoal at $l=0\text{ }\si{\meter}$ the subgoal agent causes the motion agent to stop and maintain the current robot position.

\subsubsection{Reward} 
% DONE
With the reward function, we encourage the subgoal agent to follow the planned A* path without any collisions.
It consists of three parts: 
%A penalty for collisions, a penalty for large deviations from the global A* path, and a penalty if the distance to the closest obstacle is less than a given safety distance:
\begin{equation}
	\begin{aligned}
		\label{eq:subgoal_reward} 
		r_{\text{\textit{SA}}} = r_{\mathit{collision}}+r_{\mathit{A^*}}+r_{\mathit{safety}}
	\end{aligned}
\end{equation}

Upon the robot's collision with a static or dynamic obstacle, we penalize with \mbox{$r_{\textit{collision}}=-10$}.
Since the global A* path is the optimal path to the final target position, we penalize deviations from the global A* path with \mbox{$r_{\mathit{A^*}}= -0.5 * d_{\mathit{A^*}}$}.
Here, $d_{\mathit{A^*}}$ denotes the robot's current distance to a second-next waypoint position on the A* path.
In other words, we encourage the subgoal agent to navigate back to the A* path after avoiding moving obstacles.
Thirdly, to encourage safe distance keeping to all surrounding obstacles, we linearly penalize when the distance to the closest obstacle is below a safety distance \mbox{$r_{\mathit{safety}}= -2*(0.5- d_{\mathit{closest}})$} if \mbox{$d_{\mathit{closest}} \leq 0.5 \text{ }\si{\meter}$}.
In all other cases, the rewards are 0.

%\begin{equation}
%	\label{eq:subgoal_reward_collision} 
%	r_{\textit{collision}}=
%	\begin{cases}
%		-10 &\quad \text{if collision} \\
%		0  & \quad \text{else}
%	\end{cases}
%\end{equation}
%
%Since the global A* path is the optimal path to the final target position, we penalize deviations  from the global A* path.
%In other words, we encourage the subgoal agent to navigate back to the A* path after avoiding moving obstacles:
%\begin{equation}
%	\label{eq:subgoal_reward_shortgoal} 
%	r_{\mathit{A^*}}= -0.5 * d_{\mathit{A^*}}
%\end{equation}
%Here, $d_{\mathit{A^*}}$ denotes the robot's current distance to a second-next waypoint position on the A* path.
%
%Thirdly, to encourage safe distance keeping to all surrounding obstacles, we linearly penalize when the distance to the closest obstacle is below a safety distance:
%\begin{equation}
%	\label{eq:subgoal_reward_close_dist} 
%	r_{\mathit{safety}}= 
%	\begin{cases}
%		-2*(0.5- d_{\mathit{closest}}) \quad &\text{if } d_{\mathit{closest}} \leq 0.5 \text{ }\si{\meter}  \\
%		0 \quad &\text{else}
%	\end{cases}
%\end{equation}

\subsubsection{Network Architecture} 
% DONE
Fig.~\ref{fig:architecture}c) shows the overall structure of the subgoal agent.
We employ both a \textit{lidar module} to extract relevant obstacle information and a \textit{path module} to learn the guidance from the waypoints on the global A* path.
Finally, an \textit{output module} reasons between obstacle avoidance and goal-directed behavior from the concatenated features.

Fig.~\ref{fig:architecture}d) shows the lidar module in detail, which processes the lidar data as mentioned above, see Sec.~\ref{sec:subgoal_state_space}.
At this stage, the \textit{embedding module} encodes each down-sampled lidar sector individually.
Furthermore, the feature extraction for each sector is enhanced by handing the embedding to a \textit{feature module}.
In a parallel stream, the \textit{score module} reasons about the importance of each feature via its embedding, thus calculates the attention score.
A softmax operation yields the final attention weight.
Based on the computed features and attention weights, we further compute a weighted sum of individual features.
We obtain an efficient encoding of the lidar observations.
In other words, the \textit{lidar module} has the capability of attending to specific sectors of the 2d lidar observation.
Note that there is only one embedding, feature and score module encode each sector with the same parameters, respectively.
The interplay of the features and scores of the individual lidar sectors in the weighted sum corresponds to the attention mechanism~\cite{chen_crowd-robot_2019, shi_enhanced_2022}.

The outputs of the \textit{path module} and the \textit{lidar module} are concatenated, and finally passed to an \textit{action module}.
Ultimately, this network decides upon a next subgoal position $a_{\textit{SA}} = (l, \theta)$.

All these models ar constructed as ReLU-activated fully connected feed-forward networks\footnote{Layer-wise network units: \text{Embedding module}: $\left[512, 256, 128\right]$. \text{Feature module}: $\left[256, 128, 64\right]$. \text{Score module}: $\left[128, 64, 1\right]$. \textit{Path module} $\left[128, 64, 32\right]$. \textit{Output module}: $\left[128, 64, 64\right]$. Note that the output module is completed with another non-activated layer of two or one unit, depending on the use for an actor or critic, respectively.}.

As we use the DDPG~\cite{lillicrap_continuous_2019} algorithm (see Sec.~\ref{sec:training}) that yields an actor-critic structure, the critic shares the same architecture described above with the actor.
However, one difference lies in the path module, which additionally takes the predicted subgoal position (action $a_{\text{\textit{SA}}}$) as input for the critic.

\subsection{Motion Agent}
% DONE
The motion agent's (MA) objective is to move fast, smoothly and efficiently to any given nearby subgoal position.
It does not need to consider obstacles, as we design the subgoal agent to deal with the collision avoidance.
With a fixed control frequency, the motion agent operates at 20\text{ }\si{\hertz}, which is equivalent to a time step between subsequent states of $\Delta t_{\mathit{MA}} = 0.05 \text{ }\si{\second}$.

\subsubsection{State Space} 
% DONE
The state space of the low-level motion agent must therefore only contain information about the current subgoal position $a_{\text{\textit{SA}}} = (l,\theta)$, being the action of the top level subgoal agent.
It is converted to robot-centric Cartesian coordinates $(p_x, p_y)$.
To achieve smoother driving behavior, we found it beneficial to also include the time step's velocity commands, denoted as $v^*$ and $\omega^*$ so that the full state space becomes 
\begin{eqnarray}
	\label{eq:motion_obs}
	s_{\mathit{MA}} = (v^*,\omega^*, p_x,p_y,\theta_{\mathit{diff}})
\end{eqnarray}   
Note the partially redundant representation of the subgoal both in Cartesian $(p_x,p_y)$ and polar $\theta$ coordinates, which we empirically found to improve the MA's performance.

\subsubsection{Action Space} 
% DONE
The actions of the motion agent consist of the linear velocity~$v \in [0,0.5] \text{ }\si{\meter\per\second}$and angular velocity~$ \omega \in [-\frac{\pi}{2},\frac{\pi}{2}] \text{ }\si{\radian\per\second}$: 
\begin{eqnarray}
	\label{eq:motion_action}
	a_{\mathit{MA}} = (v,\omega)
\end{eqnarray}   

\subsubsection{Reward} 
% DONE
The reward function of the motion agent encourages the robot to reach the subgoal position as quickly as possible.
This behavior can simply be encoded with a large goal reaching reward, and a distance-to-subgoal penalty: 
\begin{equation}
	\begin{aligned}
		\label{eq:motion_reward} 
		r_{\mathit{MA}} = r_{\mathit{reach}}+r_{\mathit{dist}}
	\end{aligned}
\end{equation}

with $r_{\mathit{reach}}=2$ if the subgoal is reached, else zero, and $r_{\mathit{dist}}= -1 * d_{\mathit{SG}}$, where $d_{\mathit{SG}}$ is the Euclidean distance between robot and subgoal.
%\begin{equation}
%	\label{eq:motion_reward_reach} 
%	r_{\mathit{reach}}=
%	\begin{cases}
%		2 &\quad \text{if subgoal reached} \\
%		0  & \quad \text{else}
%	\end{cases}
%\end{equation}
%
%and
%\begin{equation}
%	\label{eq:motion_reward_speed} 
%	r_{\mathit{dist}}= -1 * d_{\mathit{SG}}
%\end{equation}
%where $d_{\mathit{SG}}$ is the Euclidean distance between robot and subgoal.

\subsubsection{Network Architecture} 
% DONE
A fully connected feed-forward network is employed for the motion agent, with ReLU-activated layer units $[256,128,64,64]$.
The actor and critic in TD3~\cite{fujimoto_addressing_2018} (see Sec.~\ref{sec:training}) share the same structure of a fully connected network.

\subsection{Training of the Agents}
\label{sec:training}
% DONE
Firstly, we train the motion agent to enable the robot to reach any given nearby position.
So eventually, it will function as the motion controller for the robot.
In a simple, empty environment, we sample nearby positions in the range of $(0, 0.7]\text{ }\si{\meter}$ (probability $p=0.2$), either in a straight line, as a curvy line with direction changes of $[-\frac{\pi}{2}, +\frac{\pi}{2}]$~($p=0.3$), or with fully random direction changes ($p=0.5$).
The convergence criteria of the motion agent training are 50 consecutive successful episodes.
After the training of the motion agent is complete, we train the subgoal agent.
In this second training process, the actual collision avoidance is learned.

Both the subgoal- and motion agent are trained separately.
We have experimented with simultaneous training of both agents, which has resulted in inferior overall performance.
Additionally, the motion agent is not designed to handle collision avoidance, but to efficiently reach the predicted subgoal.
Therefore, it does not have access to lidar data but only to the subgoal position.
Hence during training, the subgoal agent learns to play a consistent and well pre-trained motion agent that does not alter its behavior over the course of training.

Regarding possible learning algorithms, we choose TD3~\cite{fujimoto_addressing_2018} for the training of the motion agent.
We found its output velocities to be more stable over time than the output of DDPG~\cite{lillicrap_continuous_2019}, resulting in smoother driving behavior of the robot.
For the training of the subgoal agent, we found DDPG~\cite{lillicrap_continuous_2019} to converge the fastest while achieving better collision avoidance performance.
In both cases, the implementations of \textit{ChainerRL} for Pytorch \cite{fujita_chainerrl_2022} are used.

\subsection{Environment Setup}
\begin{figure}[t]
	\centering
	\includegraphics[width=1.0\linewidth]{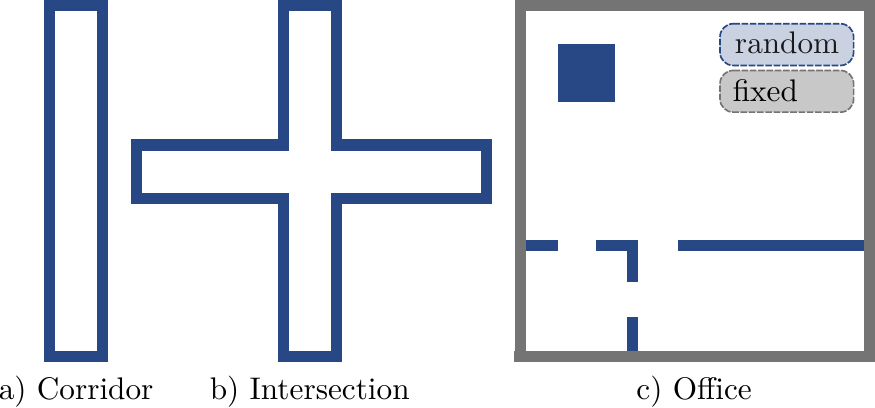}
	\caption{
		% DONE
		Three different scene types are sampled during training. 
		\textbf{a)} In the corridor and  \textbf{b)} intersection environment, the wall distances are randomized (blue). 
		\textbf{c)} In the office environment, the outer walls are fixed, but the inner wall placement is randomized for diverse room setups.
	}
	\label{fig:environments}
\end{figure}

In the following, by the term episode we refer to one full navigation run of the robot from start until a termination criteria is met: global goal reached, collision, or timeout.

\subsubsection{Scenes} 
\label{sec:scenes}
% DONE
This section highlights the environment and obstacle simulation.
Our simulation environment is based on Pybullet~\cite{coumans_pybullet_2016}.
We use three different classes of environments, a corridor scene, an intersection scene, and an office scene, see Fig.~\ref{fig:environments}.
To achieve good generalization and a smooth sim-to-real transfer, the environments need to be randomized.
We do so by randomizing the wall placements.
The corridor scenes (Fig.~\ref{fig:environments}a) have a width and length ranging from $[1.8 \text{ }\si{\meter},3 \text{ }\si{\meter}]$ to $[10 \text{ }\si{\meter}, 14 \text{ }\si{\meter}]$, respectively.
For the intersection scenes (Fig.~\ref{fig:environments}b), the hallway width and length sample in the range of $[1.8 \text{ }\si{\meter},2.5 \text{ }\si{\meter}]$ and $[4 \text{ }\si{\meter},6 \text{ }\si{\meter}]$, respectively.
The office scenes (Fig.~\ref{fig:environments}c) have a fixed outer width and length of 7 meters, but the inner wall placement is randomized so it creates different kinds and sizes of rooms in each episode.

\subsubsection{Robot} 
% DONE
The start and goal position of the robot are sampled at the beginning of each episode.
In both the corridor and intersection scene, start and goal position are sampled in different dead-ends of the scene.
For the office scene, start and goal pose are sampled in opposing corners of the outer walls.

\subsubsection{Obstacles} 
% DONE
In simulation, the obstacles are represented by cuboids.
We assume this to be a good enough representation of pedestrians, as the lidar observation is down-sampled to a low resolution anyways, see Sec.~\ref{sec:subgoal_state_space}.
For training, we use two dynamic and one static unknown obstacle, which are not contained in the given map of the environment.
During one episode, each dynamic obstacle repeatedly moves back and forth its own A* path between their sampled start and end positions.
One of the dynamic obstacle moves on the same global path as the robot but either from the middle or end to the start position of the robot's global A* path.
The other one is sampled so that it crosses the robot's global path.
A random forward speed between $[0.1, 0.5]\text{ }\si{\meter\per\second}$ is sampled before each episode for each dynamic obstacle.
The unknown static obstacle is placed randomly on the robot's planned path.
%We trained the subgoal agent with or without replanning the global path.

%For testing, we run 1,000 episodes for each method in corridor, intersection, and office scenes, respectively. We evaluated the performance on different numbers of static and dynamic objects, and also different velocities to evaluate the collision avoidance performance.

\section{Experiments}
Our experiments are designed to show the robustness and generalization capabilities of our approach both in simulation and on the real robot.

\subsection{Baseline}
\label{sec:baseline}
% DONE
We choose the approach by \mbox{P{\'e}rez-D’Arpino\etal\cite{perez2021robot}} as our baseline to demonstrate the superiority of our attention mechanism.
The authors employ a similar setup for their state space, which is lidar measurements, waypoints extracted from a global path planned and the global goal position.
However, we do not pass the global goal position directly to the subgoal agent.
The key difference is that our framework uses the attention mechanism on the lidar data, where P{\'e}rez-D’Arpino\etal apply multiple 1D convolutional layers to process lidar data.
Furthermore, a single-layer agent directly drives the robot based on the given observations, in comparison with our hierarchical approach.
We adapted their convolutional network architecture to work with our environments and learning setup.
We will demonstrate that our subgoal-driven agent based on attention outperforms the 1D convolutional lidar-processing approach by \mbox{P{\'e}rez-D’Arpino\etal\cite{perez2021robot}}.

% ===================================================================================
% Differences our implementation of the Perez-D'Arpino vs. described in their paper
% ===================================================================================
% - MLP of Goal is smaller (implemented: [32, 16], theirs: [256])
% - MLP of WP is smaller (implemented: [128, 64], theirs: [256])
% - 1D Conv channels (implemented: [32, 16, 8], theirs: unknown)
% - action timestep (implemented: unknown, ours: 0.2 s (SA), 0.05 s (MA), theirs: 0.25 s)
% - RL method (implemented: unknown, theirs: SAC)
% - WP resolution (implemented: 0.3 m, theirs: 1 m)
% - WP number (implemented: 5, theirs: 6)
%
% Issues that I see:
% - Parameters not correctly adapted from their approach
% - They do collision avoidance with up to 16 dyn. obstacles.
% - Looking at their paper, it feels like our adaption of their approach underperforms significantly

\subsection{Comparative Evaluation}
% DONE
\begin{table}[t]
	\centering
	\begin{tabular}{lcc}
		\textbf{Performance [\%]} & Success & Collision \\
		\hline
		Ours & \textbf{90.7} & \textbf{9.3}\\
		\hline
		Ours (no replan) & 72.3& 27.7\\
		\hline
		P{\'e}rez-D’Arpino\etal\cite{perez2021robot} & 78.1 & 21.9\\
		\hline
	\end{tabular}
	\caption{
		Performance comparison in corridor, intersection, and office scenes with two dynamic obstacles moving at $0.3 \text{ }\si{\meter\per\second}$ and one unknown static obstacle over 1,000 episodes.
		"Ours" refers to our standard approach with replanning of the global A* path every three subgoal predictions during navigation, while "Ours (no replan)" refers to our approach without replanning the A* path. 
		Note that no timeouts occurred, so no such data is included. 
	}
	\label{tab:performance}
\end{table}

Table~\ref{tab:performance} shows a comparison of performance between three approaches over 1,000 episodes: The main approach presented ("Ours"), an ablation study without replanning of the A* path ("Ours (no replan)"), and the baseline approach~\cite{perez2021robot}.
All scene classes were randomly sampled as described in \secref{sec:scenes}.
Two dynamic obstacles with a velocity of $0.3 \text{ }\si{\meter\per\second}$ and one unknown static object is placed in the environment.
As can be seen, our method with replanning achieves the best performance with a significantly higher success rate and a lower collision rate.
It can be concluded that the replanning plays an important role in our framework.

\begin{figure}[t]
	\centering
	\includegraphics[width=1.0\linewidth]{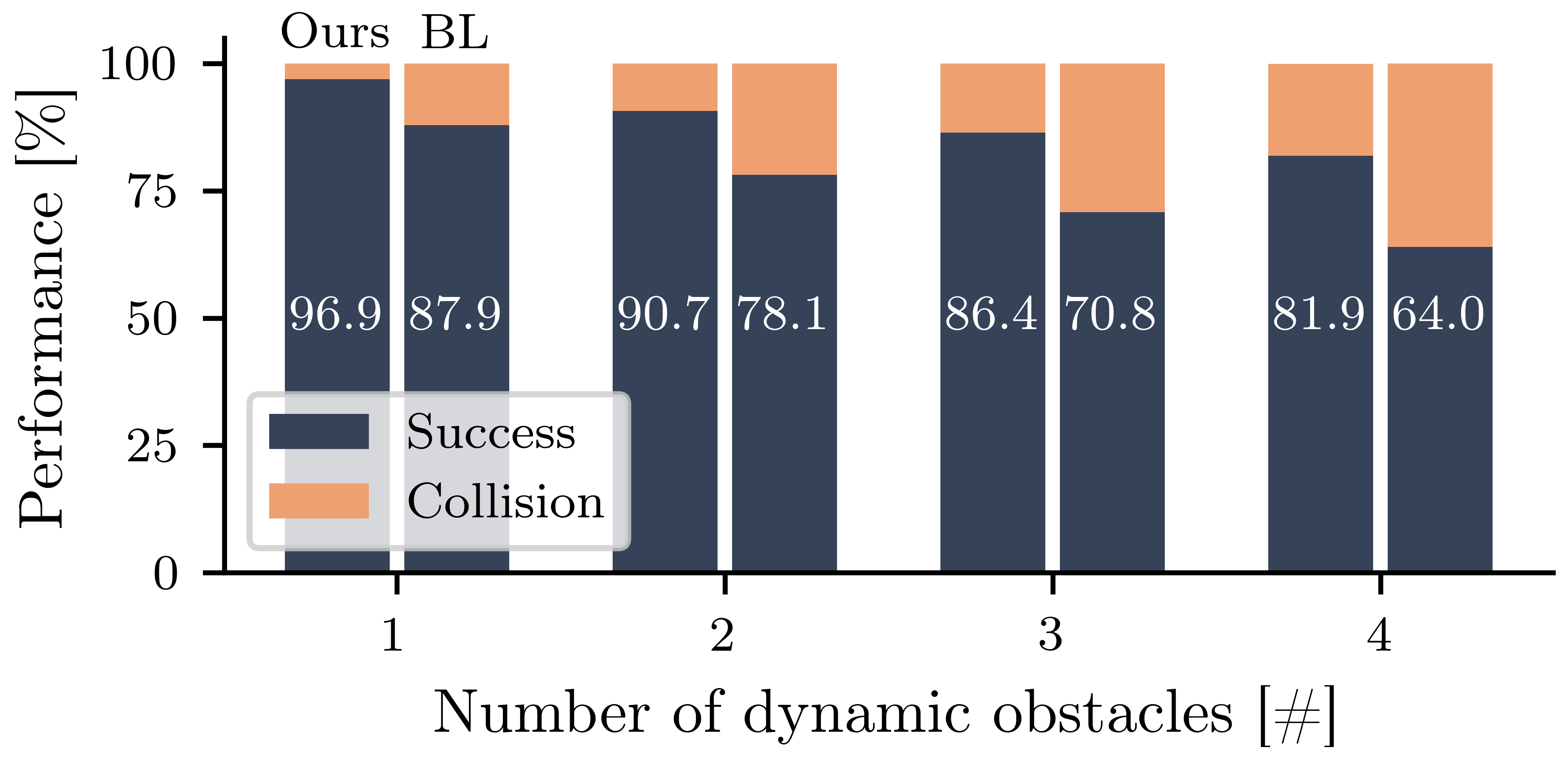}
	\caption{
		% DONE
		Performance comparison against increasing number of dynamic obstacles over 1,000 episodes
		"BL" refers to the baseline approach~\cite{perez2021robot}.
		All dynamic obstacles move at $0.3 \text{ }\si{\meter\per\second}$. 
	}
	\label{fig:dynamic_num}
      \end{figure}

Furthermore, we evaluated the performance of our method with replanning against different numbers of obstacles in the scene, see Fig.~\ref{fig:dynamic_num}.
Our method achieves an even higher success rate for one single dynamic obstacle, but consistently looses performance as the number of dynamic obstacles increases.
In all cases, the baseline approach is outperformed.

\begin{figure}[t]
	\centering
	\includegraphics[width=1.0\linewidth]{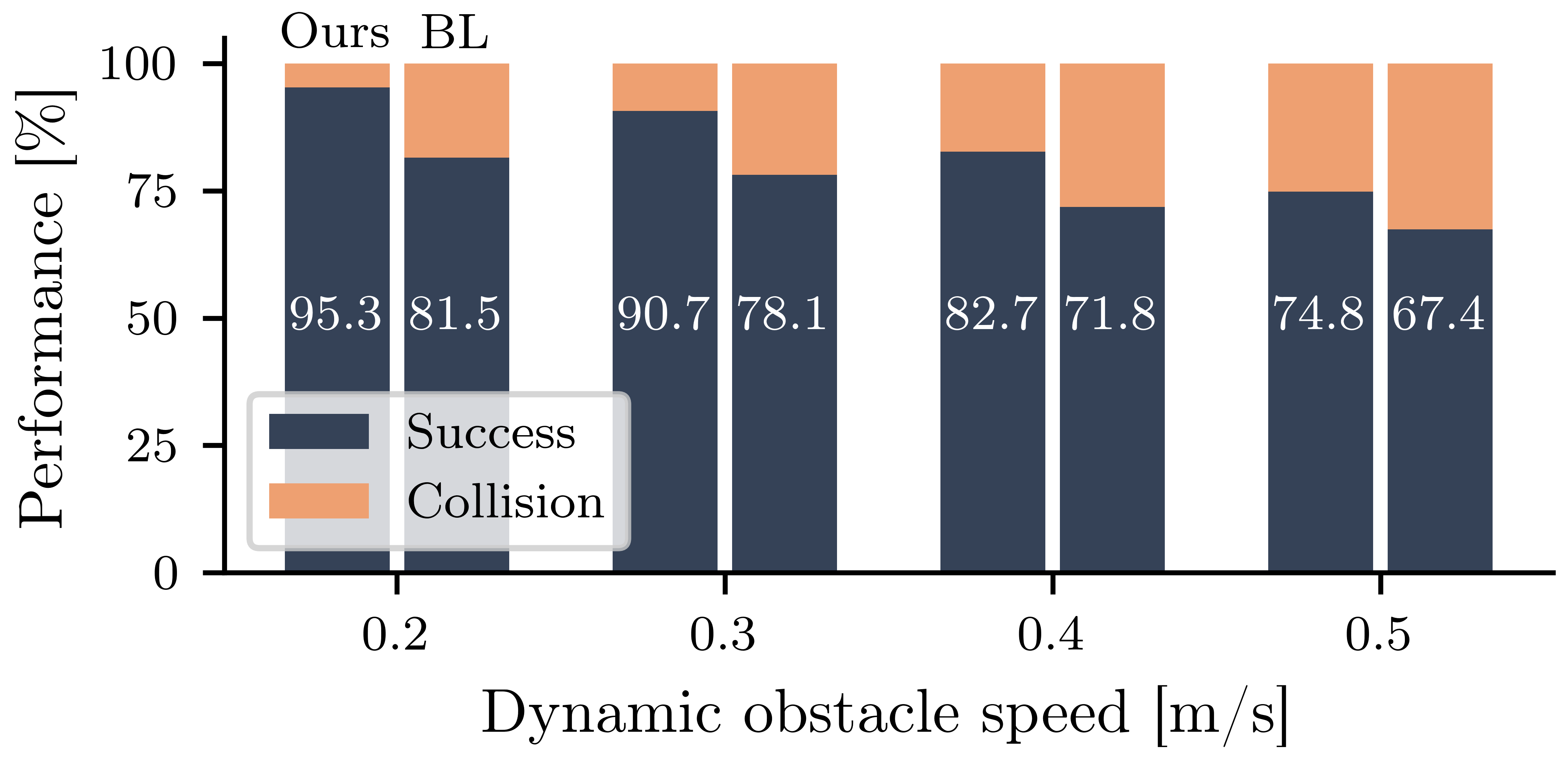}
	\caption{
		% DONE
		Performance comparison against increasing dynamic obstacle velocity over 1,000 episodes.
		"BL" refers to the baseline approach~\cite{perez2021robot}.
		All test environments contain one static obstacle on the robot's path and two dynamic obstacles. 
	}
	\label{fig:dynamic_velocity}
\end{figure}

Similarly, Fig.~\ref{fig:dynamic_velocity} shows a performance comparison under varying obstacle velocities.
Even though a deterioration can be observed for increased obstacle speeds, we measure increased performance in comparison to the baseline \cite{perez2021robot}.
The results point towards the superiority of attention mechanisms over convolutional processing of the lidar observations.
Also, decoupling obstacle avoidance and motor control may represent an advantage.

\subsection{Distribution of Subgoals}

\begin{figure}[t]
	\centering
	\includegraphics[width=1.0\linewidth]{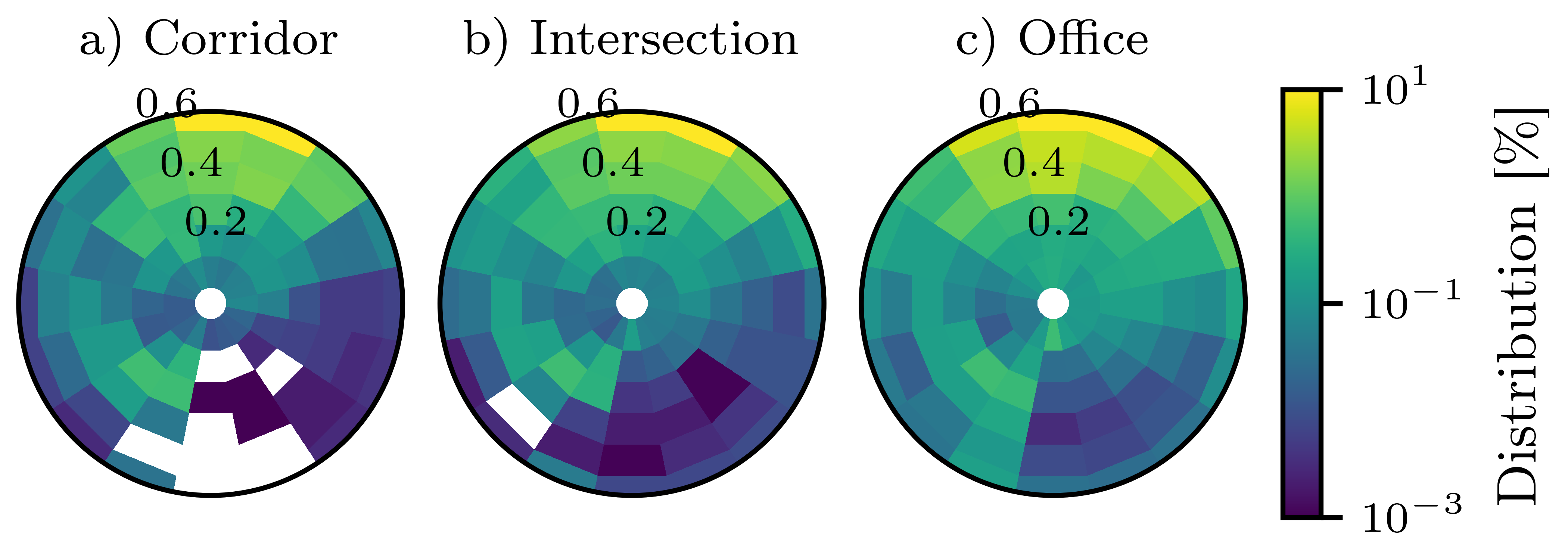}
	\caption{
		Distribution of the subgoals generated by the subgoal agent, where direction north of the plots corresponds to the robot orientation.
		A more omnidirectional distribution of subgoals can be observed in the office scene, as compared to the more straight corridor and intersection scenes.
		This translates to more directional changes.
		However, the majority of subgoals are sampled in front of the robot for all scenes.
	}
		\label{fig:action_analysis:our}
\end{figure}

Fig.~\ref{fig:action_analysis:our} shows the distribution of selected subgoals from our subgoal agent in the corridor, intersection, and office scenes.
As can be seen, most subgoals fall into the angle range of $\pm\frac{\pi}{4}$ in front of the robot.
This is reasonable, since in most cases when encountering obstacles, the robot only needs to slightly turn left or right to avoid a collision.
The distance distribution of subgoals, however, depends on the environment.
In simpler environments such as the corridor or intersection scenes, our agent prefers to move forward fast.
This also reflects the nature of the scenes being mostly straight.
In the more complex office environment, our subgoal agent behaves more conservatively through closer subgoals.
Also, more subgoals are placed behind the robot, reflecting the necessity for direction changes, e.g., to let a pedestrian pass through a door before continuing.

%\begin{figure}[t]
%	\centering
%	\subfigure[Corridor]{
%		\begin{minipage}[t]{0.31\linewidth}
%			\centering
%			\label{fig:experi:subgoal:action_analysis:corridor:lidar_attention}
%			\includegraphics[width=\linewidth]{action_corridor_c2_s1_speed0.3.png}
%		\end{minipage}%	
%	}
%	\subfigure[Intersection]{
%	\begin{minipage}[t]{0.31\linewidth}
%		\centering
%		\label{fig:experi:subgoal:action_analysis:intersection:lidar_attention}
%		\includegraphics[width=\linewidth]{action_cross_c2_s1_speed0.3.png}
%	\end{minipage}%	
%	}
%	\subfigure[Office]{
%	\begin{minipage}[t]{0.31\linewidth}
%		\centering
%		\label{fig:experi:subgoal:action_analysis:office:lidar_attention}
%		\includegraphics[width=\linewidth]{action_office_c2_s1_speed0.3.png}
%	\end{minipage}%	
%	}
%	\caption{
%		Distribution of the subgoals generated by the subgoal agent (yellow $\rightarrow$ more, blue $\rightarrow$ less).
%		A more omnidirectional distribution of subgoals can be observed in the office scene, as compared to the more straight corridor and intersection scenes.
%		This translates to more directional changes.
%		However, the majority of subgoals are sampled in front of the robot for all scenes.
%	}
%	\label{fig:action_analysis:our}
%\end{figure}

\subsection{Real-Robot Experiment}
% DONE
We also successfully tested our approach with a real robot.
To do so, we used ROS~\cite{quigley_ros_2009} to run our trained framework with the real Turtlebot.
For mapping of and localization in the environment, we relied on the ROS packages Gmapping~\cite{gmapping} and AMCL\cite{fox_monte_1999}, respectively.
In the accompanying video$^1$ we demonstrate that the learned policy transfers smoothly without noticeable sim-to-real deterioration and leads to smooth and safe trajectories around moving humans.

\section{Conclusions}
% DONE
\label{sec:concl}
In this work, we propose a novel subgoal-driven navigation architecture for robot navigation in indoor environments containing static as well as dynamic unknown obstacles.
We separate the navigation problem in two layers focusing on obstacle avoidance and velocity control respectively, and apply the attention mechanism to weigh the importance of lidar data spatially.
The experimental evaluation in simulation demonstrates that in the given environments our attention-based approach significantly outperforms a state-of-the-art baseline~\cite{perez2021robot} that processes lidar with convolutional networks with respect to success rate.
Furthermore, we showed in real-world experiments, that our agent can be transferred to a real Turtlebot robot while achieving collision-free navigation in challenging scenes.

%\todo{Check citations, arxiv format.} 

\bibliographystyle{IEEEtran}
\bibliography{refs,refs_zotero}

\end{document}